%% file: main.tex
\documentclass[sigconf]{acmart}
\usepackage[utf8]{inputenc} 
\usepackage[T1]{fontenc}    
\usepackage{hyperref}       
\usepackage{url}            
\usepackage{booktabs}       
\usepackage{amsfonts}       
\usepackage{nicefrac}       
\usepackage{microtype}      
\usepackage{xcolor}         
\usepackage{graphicx}       

\usepackage{multirow} 
\usepackage{algorithm,algorithmic}

\newlength{\textfloatsepsave} 
\AtBeginDocument{%
  \providecommand\BibTeX{{%
    \normalfont B\kern-0.5em{\scshape i\kern-0.25em b}\kern-0.8em\TeX}}}

\setcopyright{none}
\renewcommand\footnotetextcopyrightpermission[1]{}
\acmConference[KDD '22 DSHealth Workshop]{}{August 14, 2022}{Washington DC}

\begin{document}

\title{FedPseudo: Pseudo value-based Deep Learning Models for Federated Survival Analysis}

\author{Md Mahmudur Rahman}
\email{mrahman6@umbc.edu}
\affiliation{%
  \institution{University of Maryland, Baltimore County}
  \city{Baltimore}
  \state{Maryland}
  \country{USA}
}

\author{Sanjay Purushotham}
\email{psanjay@umbc.edu}
\affiliation{%
  \institution{University of Maryland, Baltimore County}
  \city{Baltimore}
  \state{Maryland}
  \country{USA}
}

\renewcommand{\shortauthors}{Md Mahmudur Rahman and Sanjay Purushotham}
\begin{abstract}
Survival analysis, time-to-event analysis, is an important problem in healthcare since it has a wide-ranging impact on patients and palliative care. Many survival analysis methods have assumed that the survival data is centrally available either from one medical center or by data sharing from multi-centers. However, the sensitivity of the patient attributes and the strict privacy laws have increasingly forbidden sharing of healthcare data. To address this challenge, the research community has looked at the solution of decentralized training and sharing of model parameters using the Federated Learning (FL) paradigm. In this paper, we study the utilization of FL for performing survival analysis on distributed healthcare datasets. Recently, the popular Cox proportional hazard (CPH) models have been adapted for FL settings; however, due to its linearity and proportional hazards assumptions, CPH models result in suboptimal performance, especially for non-linear, non-iid, and heavily censored survival datasets. To overcome the challenges of existing federated survival analysis methods, we leverage the predictive accuracy of the deep learning models and the power of pseudo values to propose a first-of-its-kind, pseudo value-based deep learning
model for federated survival analysis (FSA) called \textbf{FedPseudo}. Furthermore, we introduce a novel approach of deriving pseudo values for survival probability in the FL settings that speeds up the computation of pseudo values.
Extensive experiments on synthetic and real-world datasets show that our pseudo valued-based FL framework achieves similar performance as the best centrally trained deep survival analysis model. Moreover, our proposed FL approach obtains the best results for various censoring settings.
\end{abstract}



\maketitle
\section{Introduction}
\input{Introduction.tex}

\vspace{-10pt}
\section{Background}
\input{Background.tex}
\vspace{-3pt}
\section{Our Proposed Method: FedPseudo}
\input{Method.tex}

\vspace{-5pt}
\section{Experiments}
\input{Experiments.tex}

 \vspace{-10pt}
\section{Results and Discussion}
\input{Real_Result.tex}

\vspace{-10pt}
\section{Conclusion}
\input{Conclusion.tex}

\section*{Acknowledgement}
This work is supported by grant IIS–1948399 from the US National Science Foundation (NSF).

\bibliographystyle{ACM-Reference-Format}
\bibliography{references}

\end{document}

%% file: Introduction.tex
Survival analysis, time-to-event analysis, is one of the important problems studied in healthcare. 
Solving it has a wide-ranging impact, including better patient and palliative care, treatment and therapy interventions, and hospital resource allocations, to name a few.
Modern machine learning algorithms such as deep learning models have achieved excellent performance for survival analysis~\citep{ katzman2018deepsurv, zhao2020deep, rahman2021deeppseudo}. Most of the research and model development in survival analysis has assumed that the survival data is centrally available either from one medical center or by data sharing from multi-centers. However, the sensitivity of the patient data and strict privacy laws and regulations forbid the sharing of data, which results in training challenges for the machine learning models as they require a large amount of training data to learn model parameters. Therefore, there is a need to develop survival analysis algorithms to learn from distributed survival datasets. Federated learning (FL), an emerging distributed machine learning paradigm, has become popular for learning from decentralized datasets and could be a promising solution for performing survival analysis from multi-center datasets. However, multi-center healthcare data, in general, can be heterogeneous, non-iid, and non-uniformly censored. These challenges can pose additional hurdles for performing FL for survival analysis. The main goal of this paper is to develop a new class of pseudo valued-based deep learning approaches for survival analysis in a Federated Learning framework.  

Recently, Federated learning has been used in many applications, including healthcare~\citep{li2021federated}. 
Even though survival analysis has been extensively studied in statistics \citep{cox1972regression}, machine learning and healthcare informatics communities \citep{rahman2021deeppseudo, katzman2018deepsurv, kvamme2019time,zhao2020deep,lee2018deephit}, there are only a few works on applying FL for survival analysis \citep{andreux2020federated,wang2022survmaximin}.
Andreux et al.~\citep{andreux2020federated} proposed a discrete-time extension of the Cox proportional hazard model by formulating the survival analysis as a classification problem with a separable objective function. However, using the CoxPH model in the FL framework needs to satisfy strong underlying assumptions, such as linearity, proportional hazards, and common baseline hazard function among clients, which are often violated in the real-world survival datasets. 
A common baseline hazard function is difficult to obtain as different clients have different baseline hazard functions, and proportional hazard assumption is difficult to hold in every center, and thus, the CoxPH-based FL framework for survival analysis might lead to suboptimal solutions compared to centralized survival models. 

To address the limitations of the existing FL approaches for survival analysis, we propose a first-of-its-kind, pseudo value-based deep learning models for federated survival analysis (FSA) called \textbf{FedPseudo}. FedPseudo relaxes the strong underlying assumptions needed in the existing FSA approaches and leverages the predictive accuracy of the deep learning models and the power of pseudo values in handling censoring to obtain accurate survival analysis results in a federated setup. Along with the FedPseudo, we introduce a novel and fast approach of deriving pseudo values for survival probability for the FL settings, which allows us to model and analyze the survival data from different clients without sharing data. Extensive experiments on both synthetic datasets and real-world survival datasets show that our FedPseudo obtains similar performance as the best centrally trained deep survival models (gold standard) and performs similar to or better than the existing FSA approaches, especially for various censoring settings.

%% file: Background.tex
\textbf{Survival Data:}
Let’s consider a survival dataset $D$ comprising of time-to-event information about $N$ subjects who are followed up for a ﬁnite amount of time. For a subject i, $D$ is a tuple, $D = \{X_i, Y_i, \delta_i\}$ , where $i=1,2,…,N$ . $X_i \in R^{p}$ is a p-dimensional vector of covariates for the subject $i$ and $Y_i$ is the nonnegative observed time for subject $i$. Let, $T_i$ is the nonnegative survival time and $C_i$ is the nonnegative censoring time for subject $i$. Then the observed time for subject $i$, $Y_i=min(T_i, C_i)$ and the event indicator $\delta_i=\mathbb{I}(T_i\leq C_i)$. In other words, $\delta_i=1$, if the $i^{th}$ subject's observed time $Y_i$ is the event time, $T_i$, whereas $\delta_i=0$, if the $i^{th}$ subject's event time is unknown and we only know the observed time of subject $i$ is greater than the censoring time $C_i$. Our goal is to estimate the conditional survival function $S(t|\mathbf{X})=P(T>t|\mathbf{X}=x)$ at a particular time $t$ given the covariates $\mathbf{X}$ for a subject in a client. 

\textbf{Federated Learning:} The idea of federated learning (FL) was first introduced by Google \cite{mcmahan2017communication}. In the first step of FL, an aggregation server initialize a global model $w^{0}$ and set the hyperparameters, such as learning rate, batch size, and the number of iterations for training. The global model is sent to the $k$ number of clients. At each communication round $v$, each of the local clients update their local models $w^{v}_{k}$ based on the global model $w^{v}$. The updated local models $w^{v}_{k}$ are sent back to the aggregation/global server after completing each communication round. The aggregation/global server averages the local updated model and generate a new global updated model by $w^{v+1}=\frac{1}{K}\sum_{k=1}^{K}w^{v}_{k}$, which is sent back to the local clients for the next communication round. Federated learning algorithms overcome the problem of the limited amount of data in decentralized health centers to develop an efficient model and preserve the privacy protocol for clinical datasets.

\vspace{-5pt}

%% file: Method.tex
\begin{algorithm}[tb]
\caption{Derivation of pseudo values for the survival probability in FL setting}
\label{alg1}
\begin{algorithmic}
\REQUIRE Observed times $Y_{k}$ and event status $\delta_{k}$ of the local client $Cl_{k}$; $k=\{1, 2,..,K\}$ where $Cl$ is the set of clients, $Cl=\{Cl_1,Cl_2,..,Cl_K\}$, time horizon $T_h$, pre-specified vector of time points at which pseudo values will be calculated $\mathbf{\tau}=\{\tau_1<\tau_2<..<\tau_M\}$, number of individuals $n_{k}$ in client $Cl_k$, total number of individuals, $N=\sum_{k=1}^{K}n_{k}$.
\ENSURE Pseudo values for $i^{th}$ individual in client $k$; ${J}_{ki}(t)$
\FOR{\texttt{$k\in Cl$ in parallel}}
\STATE Compute the maximum observed time $T^{k}_{max}$
\STATE Send $T^{k}_{max}$ to the global server
\ENDFOR
\STATE $T_{minmax}\gets \min_{k\in Cl}T^{k}_{max}$
\STATE Specify $T_h=\{t_{h1},t_{h2},...,t_{hm}\}$, where  $t_{hj}\leq T_{minmax}; j={1,2,..,m}$
\FOR{\texttt{$k\in Cl$ in parallel}} 
\STATE Create a partial matrix $M_{k}$; $[r_{k1}, d_{kj}, c_{kj}]\in M_{k}$; for every $t_{hj}$, where, $r_{kj}$ is the number of subjects not experiencing a event up to the time point $t_{hj}$, $d_{kj}$ is the total number of event observed at time $t_{hj}$, and $c_{kj}$ is the total number of censored observations at time $t_{hj}$.
\STATE Send $M_{k}$ to the global server
\ENDFOR
\STATE $M\gets \sum_{k\in Cl}M_{k}$
\STATE $r^{\prime}_{1}, d^{\prime}_{j}, c^{\prime}_{j}\in M$; for every $t_{hj}$
\FOR{\texttt{$j= 2,...,m$}}
\STATE $r^{\prime}_{j}=r^{\prime}_{j-1}-d^{\prime}_{j-1}-c^{\prime}_{j-1}$
\ENDFOR
\STATE $\hat{S}(t)=\prod_{j:t_{hj}\leq t}\frac{r^{\prime}_{j}-d^{\prime}_{j}}{r^{\prime}_{j}}$
\FOR{\texttt{$k\in Cl$ in parallel}} 
\STATE Send aggregated partial matrix $M$ to client $Cl_k$
\FOR{\texttt{$i= 1, 2,...,N$}}
\STATE $r^{\prime-i}_{k1}=r^{\prime}_{1}-1$ 
\IF{$T_{ki}\leq t_{hj}$ and $\delta_{ki}=1$} 
\STATE $d^{\prime-i}_{kj} \gets d^{\prime}_{j}-1$
\ENDIF
\IF{$T_{ki}\leq t_{hj}$ and $\delta_{ki}=0$} 
\STATE $c^{\prime-i}_{kj} \gets c^{\prime}_{j}-1$
\ENDIF  
\FOR{\texttt{$j= 2,...,m$}}
\STATE $r^{\prime-i}_{kj}=r^{\prime-i}_{k,(j-1)}-d^{\prime-i}_{k,(j-1)}-c^{\prime-i}_{k,(j-1)}$
\ENDFOR
\STATE $\hat{S}_{k^{\prime}}^{-i}(t)=\prod_{j:t_{hj}\leq t}\frac{r^{\prime-i}_{kj}-d^{\prime-i}_{kj}}{r^{\prime-i}_{kj}}$
\STATE ${J_{ki}}(t)=N*\hat{S}(t)-(N-1)*\hat{S}_{k^{\prime}}^{-i}(t)$
\ENDFOR  
\ENDFOR 
\end{algorithmic}
\end{algorithm}

\begin{figure*}[h]
    \includegraphics[trim=0 38 0 0,clip,width=0.75\textwidth]{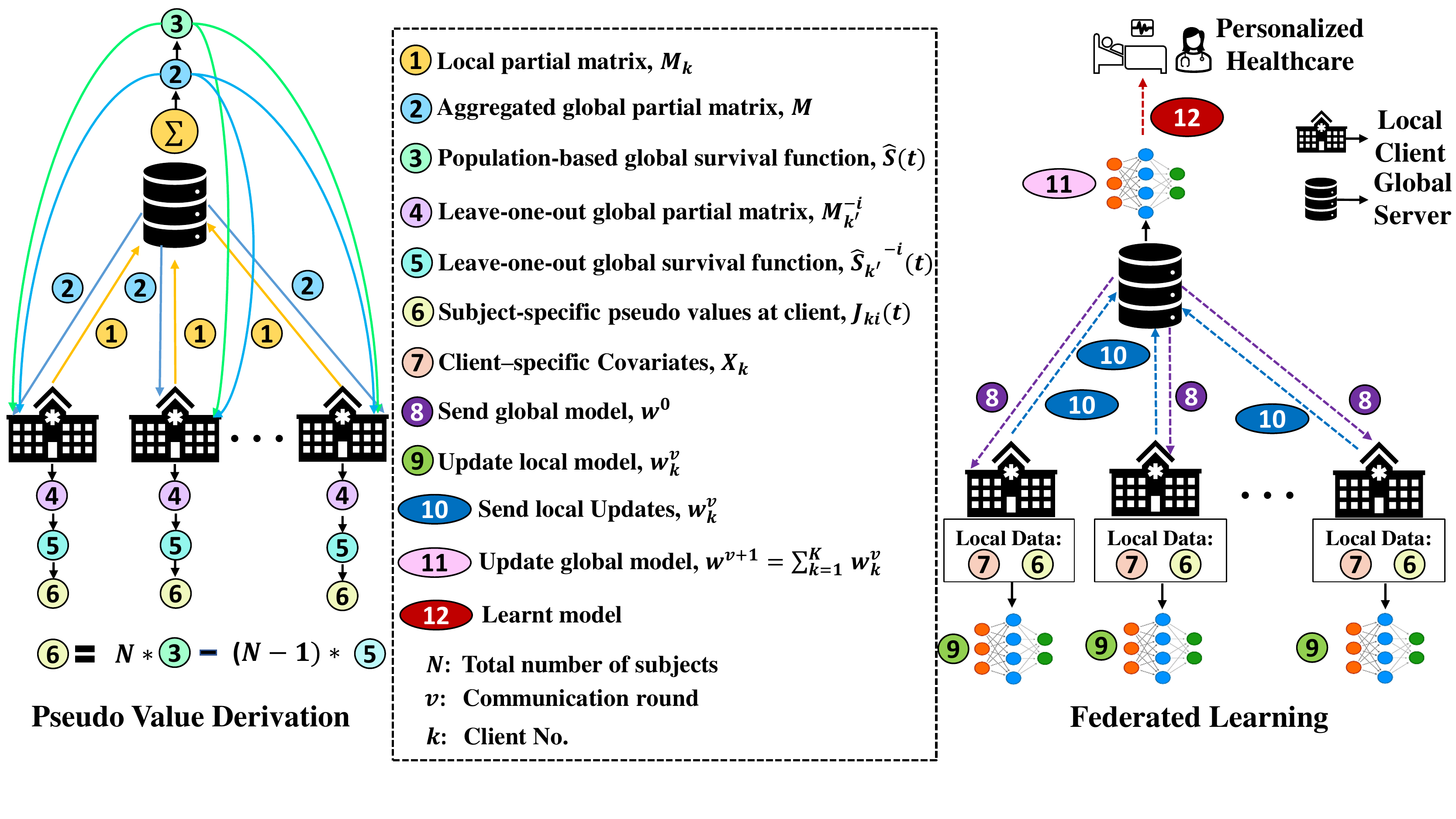}
    \vspace{-5pt}
    \caption{Our Proposed Federated Survival Analysis Approach: \textbf{FedPseudo}}
    \label{fig:FedSurv}
    \vspace{-12pt}
\end{figure*}

We propose \textbf{FedPseudo} for FSA, where we federate a number of client-specific pseudo value-based deep survival models. 
At each client, we formulate the survival analysis problem as a simple regression analysis problem 
using pseudo values for survival function as a quantitative response variable. The pseudo values are known to handle censoring and enable the model to obtain subject-specific predictions of survival probability. 
The local model takes the client's covariates, $X_k$, as input and the client's subject-specific pseudo values (${J}_{ki}(t)$) as the target variable. 
In the existing pseudo value-based deep models \citep{zhao2020deep, rahman2021deeppseudo}, the mean squared error (MSE) loss function is used for model training. 
However, the different range of the ground-truth pseudo values and predicted survival probabilities can lead the MSE to converge to poor local optima. 
Thus, we introduce a pseudo value-based loss function for client $k$ at time $t$, $L_{k}(t)$ as, $    L_{k}(t)=\frac{1}{n_{k}}\sum_{i=1}^{n_{k}}[{J}_{ki}(t)(1-2\hat{S}(t|x_ki))+\hat{S}^{2}(t|x_ki)]$, where, $\hat{S}(t|x_ki)$ and ${J}_{ki}(t)$ respectively are the predicted survival probability and the pseudo values at time point $t$ for $i^{th}$ individual in client $k$. 

In the federated settings, 
we need to carefully derive the pseudo values since the data at different clients (e.g., different health centers) may have different data distributions (non-iid). 
Moreover, pseudo values in the centralized settings are calculated on the entire sample, which is infeasible in federated and decentralized settings due to strict privacy protocols about data exchange among clients.
To overcome these challenges, we introduce a novel pseudo value derivation algorithm for FL settings, where we utilize the statistics from the local clients' data without sharing the sensitive patient data across the clients. First, local partial matrices calculated at clients are sent to the global server, where they are aggregated to obtain the global partial matrix and global survival function.
The global partial matrix is sent back to the clients to compute the leave-one-out survival function for each client. 
Then, each client computes the subject-specific pseudo values for its patients using the population-based global survival function and the leave-one-out survival function. The details of the pseudo value derivation are shown in \textbf{Algorithm 1}.

\textbf{Federated Training:} In our proposed approach, FedPseudo, a global pseudo value-based deep model, is sent to the randomly selected local clients from the global server during each communication round. Each selected client updated their local models with the global model parameter and then trained using their local data. The updated models from local clients are sent to the global server, which aggregates the local models' updates to update the global model using an FL algorithm such as FedAvg \citep{mcmahan2016communication}. The global server sends the updated global model back to the local clients, and we repeat the same procedure for some pre-specified communication rounds. The details are shown in figure \ref{fig:FedSurv}.

%% file: Experiments.tex
We conducted experiments to answer these questions:
(a) how does FedPseudo compare to the existing FSA approaches? 
(b) how well do models perform under different censoring settings?
(c) FSA models' performance compared to centralized survival models?

\begin{table}[h]
\vspace{-5pt}
\caption{Descriptive statistics of the datasets}
\vspace{-10pt}
\label{tab:des_stat}
\renewcommand{\arraystretch}{1.2}
\resizebox{0.47\textwidth}{!}{
\begin{tabular}{c|c|c|c|c|c}
\hline
\textbf{Dataset} &
  \textbf{\begin{tabular}[c]{@{}c@{}}\# features \\ (\%) \end{tabular}} &
  \textbf{\begin{tabular}[c]{@{}c@{}}\# Uncensored \\ (\%) \end{tabular}} & 
  \textbf{\begin{tabular}[c]{@{}c@{}}\# Censored \\ (\%) \end{tabular}} & 
  \textbf{\begin{tabular}[c]{@{}c@{}}\# Clients \\ (\%) \end{tabular}} & 
  \textbf{\begin{tabular}[c]{@{}c@{}}\# training \\ data/ client\end{tabular}} \\ \hline
METABRIC & 9  & 1103 (58\%)   & 801 (42\%)     & 10 & 152  \\ \hline
SUPPORT  & 14 & 6036 (68\%)   & 2837 (32\%)    & 10 & 709  \\ \hline
\begin{tabular}[c]{@{}c@{}} CC 25\%\end{tabular}    & 12 & 15000 (75\%)  & 5000 (25\%)  & 10 & 1600  \\ \hline
\begin{tabular}[c]{@{}c@{}} CC 50\%\end{tabular}    & 12 & 10000 (50\%) & 10000 (50\%) & 10 & 1600 \\ \hline
\begin{tabular}[c]{@{}c@{}} CC 75\%\end{tabular}     & 12 & 5000 (25\%) & 15000 (75\%) & 10 & 1600 \\ \hline
\end{tabular}}
\vspace{-10pt}
\end{table}
\noindent \textbf{\Large Datasets:} 
Table \ref{tab:des_stat} shows the descriptive statistics of the datasets.\\
\textbf{Real datasets:} (a) \textbf{METABRIC:} The Molecular Taxonomy of Breast Cancer International Consortium (METABRIC) dataset aims at determining new breast cancer subgroups who are at risk of death using the gene and protein expression profiles and clinical information of patients. 
(b) \textbf{SUPPORT:} Study to Understand Prognoses Preferences Outcomes and Risks of Treatment (SUPPORT) is the study of the survival time of seriously ill hospitalized adults \citep{knaus1995support}. 

\noindent We used the same train-test splits \& features as in \citep{katzman2018deepsurv} for real data.

\textbf{Synthetic Datasets with Different Censoring Settings:} We first construct a complete follow-up (uncensored) dataset by generating 12 covariates from a multivariate normal distribution with mean $\mu$ and variance $\sigma^2$ followed by Weibull distributed survival times through Cox models \citep{machado2014simulation}, taking the nonlinear combination of covariates. Then we generate 3 \textbf{case censoring (CC)} datasets by randomly selecting cases with a censoring probability 0.25, 0.50, and 0.75, respectively, from the uncensored dataset. The corresponding time for the selected cases is shortened by a random amount and the event status is updated from 1 (uncensored) to 0 (censored).

\begin{table*}[h]
\caption{Performance comparisons of the models on real-world datasets.}
\vspace{-10pt}
\label{tab:real_data}
\renewcommand{\arraystretch}{1.2}
\resizebox{0.95\textwidth}{!}{
\begin{tabular}{c|c|cccc||cccc}
\hline
\multirow{2}{*}{\textbf{Dataset}} &
  \multirow{2}{*}{\textbf{\begin{tabular}[c]{@{}c@{}}Evaluation \\ Metric\end{tabular}}} &
  \multicolumn{4}{c||}{\textbf{Centralized}} &
  \multicolumn{4}{c}{\textbf{Federated Learning}} \\ \cline{3-10} 
 &
   &
  \multicolumn{1}{c|}{\textbf{CoxPH}} &
  \multicolumn{1}{c|}{\textbf{DeepSurv}} &
  \multicolumn{1}{c|}{\textbf{DeepHit}} &
  \textbf{FedPseudo} &
  \multicolumn{1}{c|}{\textbf{CoxPH}} &
  \multicolumn{1}{c|}{\textbf{DeepSurv}} &
  \multicolumn{1}{c|}{\textbf{DeepHit}} &
  \textbf{FedPseudo} \\ \hline
\multirow{2}{*}{\textbf{METABRIC}} &
  \textbf{Cindex} &
  \multicolumn{1}{c|}{0.63 (0.004)} &
  \multicolumn{1}{c|}{0.63 (0.011)} &
  \multicolumn{1}{c|}{\textbf{0.68 (0.015)}} &
  \textbf{0.67 (0.008)} &
  \multicolumn{1}{c|}{0.63 (0.006)} &
  \multicolumn{1}{c|}{0.62 (0.011)} &
  \multicolumn{1}{c|}{0.64 (0.031)} &
  \textbf{0.65 (0.021)} \\ \cline{2-10} 
 &
  \textbf{iBrier} &
  \multicolumn{1}{c|}{0.19 (0.002)} &
  \multicolumn{1}{c|}{0.19 (0.003)} &
  \multicolumn{1}{c|}{0.20 (0.003)} &
  \textbf{0.19 (0.003)} &
  \multicolumn{1}{c|}{\textbf{0.19 (0.001)}} &
  \multicolumn{1}{c|}{0.20 (0.009)} &
  \multicolumn{1}{c|}{0.20 (0.004)} &
  0.21 (0.013) \\ \hline
\multirow{2}{*}{\textbf{SUPPORT}} &
  \textbf{Cindex} &
  \multicolumn{1}{c|}{0.60 (0.001)} &
  \multicolumn{1}{c|}{0.61 (0.003)} &
  \multicolumn{1}{c|}{0.58 (0.009)} &
  \textbf{0.62 (0.005)} &
  \multicolumn{1}{c|}{0.60 (0.003)} &
  \multicolumn{1}{c|}{0.60 (0.006)} &
  \multicolumn{1}{c|}{0.58 (0.011)} &
  \textbf{0.61 (0.008)} \\ \cline{2-10} 
 &
  \textbf{iBrier} &
  \multicolumn{1}{c|}{0.21 (0.000)} &
  \multicolumn{1}{c|}{0.21 (0.002)} &
  \multicolumn{1}{c|}{0.22 (0.002)} &
  \textbf{0.20 (0.003)} &
  \multicolumn{1}{c|}{\textbf{0.21 (0.001)}} &
  \multicolumn{1}{c|}{0.21 (0.003)} &
  \multicolumn{1}{c|}{0.23 (0.007)} &
  0.22 (0.012) \\ \hline
\end{tabular}}
\vspace{-5pt}
\end{table*}

\begin{table*}[h]
\caption{Performance comparisons of the models on synthetic datasets with different censoring settings}
\vspace{-10pt}
\label{tab:synthetic}
\renewcommand{\arraystretch}{1.2}
\resizebox{0.95\textwidth}{!}{
\begin{tabular}{c|c|cccc||cccc}
\hline
\multirow{2}{*}{\textbf{Dataset}} &
  \multirow{2}{*}{\textbf{\begin{tabular}[c]{@{}c@{}}Evaluation \\ Metric\end{tabular}}} &
  \multicolumn{4}{c||}{\textbf{Centralized}} &
  \multicolumn{4}{c}{\textbf{Federated Learning}} \\ \cline{3-10} 
 &
   &
  \multicolumn{1}{c|}{\textbf{CoxPH}} &
  \multicolumn{1}{l|}{\textbf{DeepSurv}} &
  \multicolumn{1}{l|}{\textbf{DeepHit}} &
  \textbf{FedPseudo} &
  \multicolumn{1}{c|}{\textbf{CoxPH}} &
  \multicolumn{1}{l|}{\textbf{DeepSurv}} &
  \multicolumn{1}{l|}{\textbf{DeepHit}} &
  \textbf{FedPseudo} \\ \hline
\multirow{2}{*}{\textbf{\begin{tabular}[c]{@{}c@{}}Case \\ Censoring 25\%\end{tabular}}} &
  \textbf{Cindex} &
  \multicolumn{1}{c|}{0.51 (0.005)} &
  \multicolumn{1}{c|}{\textbf{0.78 (0.002)}} &
  \multicolumn{1}{c|}{0.76 (0.002)} &
  0.77 (0.002) &
  \multicolumn{1}{c|}{0.51 (0.007)} &
  \multicolumn{1}{c|}{0.77 (0.003)} &
  \multicolumn{1}{c|}{0.74 (0.011)} &
  \textbf{0.77 (0.002)} \\ \cline{2-10} 
 &
  \textbf{iBrier} &
  \multicolumn{1}{c|}{0.21 (0.000)} &
  \multicolumn{1}{c|}{0.13 (0.001)} &
  \multicolumn{1}{c|}{0.18 (0.001)} &
  \textbf{0.13 (0.001)} &
  \multicolumn{1}{c|}{0.21 (0.000)} &
  \multicolumn{1}{c|}{0.13 (0.002)} &
  \multicolumn{1}{c|}{0.18 (0.003)} &
  \textbf{0.13 (0.001)} \\ \hline
\multirow{2}{*}{\textbf{\begin{tabular}[c]{@{}c@{}}Case \\ Censoring 50\%\end{tabular}}} &
  \textbf{Cindex} &
  \multicolumn{1}{c|}{0.51 (0.002)} &
  \multicolumn{1}{c|}{0.77 (0.002)} &
  \multicolumn{1}{c|}{0.75 (0.005)} &
  \textbf{0.77 (0.003)} &
  \multicolumn{1}{c|}{0.51 (0.003)} &
  \multicolumn{1}{c|}{0.75 (0.004)} &
  \multicolumn{1}{c|}{0.74 (0.003)} &
  \textbf{0.77 (0.003)} \\ \cline{2-10} 
 &
  \textbf{iBrier} &
  \multicolumn{1}{c|}{0.22 (0.000)} &
  \multicolumn{1}{c|}{\textbf{0.13 (0.002)}} &
  \multicolumn{1}{c|}{0.18 (0.002)} &
  0.14 (0.001) &
  \multicolumn{1}{c|}{0.22 (0.000)} &
  \multicolumn{1}{c|}{0.14 (0.001)} &
  \multicolumn{1}{c|}{0.18 (0.003)} &
  \textbf{0.14 (0.001)} \\ \hline
\multirow{2}{*}{\textbf{\begin{tabular}[c]{@{}c@{}}Case \\ Censoring 75\%\end{tabular}}} &
  \textbf{Cindex} &
  \multicolumn{1}{c|}{0.50 (0.003)} &
  \multicolumn{1}{c|}{\textbf{0.78 (0.001)}} &
  \multicolumn{1}{c|}{0.75 (0.15)} &
  0.76 (0.009) &
  \multicolumn{1}{c|}{0.50 (0.003)} &
  \multicolumn{1}{c|}{0.75 (0.012)} &
  \multicolumn{1}{c|}{0.72 (0.012)} &
  \textbf{0.76 (0.009)} \\ \cline{2-10} 
 &
  \textbf{iBrier} &
  \multicolumn{1}{c|}{0.18 (0.000)} &
  \multicolumn{1}{c|}{\textbf{0.11 (0.000)}} &
  \multicolumn{1}{c|}{0.15 (0.002)} &
  0.14 (0.003) &
  \multicolumn{1}{c|}{0.18 (0.000)} &
  \multicolumn{1}{c|}{\textbf{0.12 (0.005)}} &
  \multicolumn{1}{c|}{0.15 (0.003)} &
  0.14 (0.003) \\ \hline
\end{tabular}}
\vspace{-10pt}
\end{table*}

\noindent \textbf{\large Implementation Details:}
We assume a collaboration of 10 clients (hospitals) to train a model, which is a reasonable number of clients that would likely participate in a study of a particular disease \citep{tomczak2015cancer, linge2016hpv, rahimian2022practical}. We randomly split the datasets into 80\% training and 20\% test data and randomly distributed an equal number of observations to 10 clients. We run the experiment 5 times with different seeds and report the average value of the evaluation metrics with their corresponding standard deviation. We evaluate the performance of the models in both centralized and federated settings.\\
\textbf{Centralized:} To examine the effect of FL in survival analysis, we first need to have a gold standard performance for comparison, which can be obtained in a centralized setting, i.e., training the models on the entire training dataset at a centralized server. For centralized training, the model architecture of our FedPseudo model consists of 5 hidden layers with a number of units [128,64,64,32,32]. We train our FedPseudo model up to 1000 epochs using Adam Optimizer \citep{kingma2014adam} with an early stopping criterion based on the best validation C-Index having a patience of 50. We set the learning rate $10^{-3}$, dropout $0.1$, use \texttt{SELU} activation function in the hidden layers, and use \texttt{Sigmoid} activation function in the output layer to get the survival probabilities at the pre-specified time points ($10^{th}$ percentile to $80^{th}$ percentile of the time horizon with an interval of $10$ where the maximum time horizon is the minimum of maximum survival times of the 10 clients). For the baseline models, we follow the implementation in \citep{rahimian2022practical}. \\
\textbf{Federated:} For federated learning experiments, we assume 75\% randomly selected clients are active in each communication round. We use the same FedPseudo neural network architecture and hyperparameters described in the centralized training. We use the Vanilla FL algorithm, FedAvg \citep{mcmahan2016communication} to train our FedPseudo model in the FL setting.
We set the number of communication round $T=50$, and the selected local clients' models are trained up to 1000 epochs with an early stopping criterion based on the best validation C-Index for our model and validation loss for other models, having a patience of 50 similar to the centralized training. 

\noindent \textbf{Models Comparison: }
We compare our proposed model, {FedPseudo}, to three baseline survival models: Cox Proportional Hazard Model (CoxPH) \citep{cox1972regression}, Deep CoxPH (DeepSurv) \citep{katzman2018deepsurv}, DeepHit \citep{lee2018deephit}, in both centralized and FL settings. 
 We adopted the implementation of the baseline survival models from the \textbf{pycox}~\citep{abcde} package.

\noindent \textbf{Evaluation Metrics: }
We evaluate the models using time-dependent concordance index (Cindex) \citep{antolini2005time} and integrated IPCW Brier score (iBrier) \citep{graf1999assessment}.
We use \textbf{pycox}~\citep{abcde}
package to compute the Cindex and iBrier scores.
The higher value of C-index and lower values of iBrier indicate better performance.

%% file: Real_Result.tex
\textbf{Real-world data results:} 
Table \ref{tab:real_data} shows that our  \texttt{FedPseudo} outperforms in most of the cases in centralized setting. 
In federated settings, \texttt{FedPseudo} gives better performance with respect to Cindex and similar performance with respect to iBrier.

\noindent \textbf{Simulated data results:}
Table \ref{tab:synthetic} shows that our \texttt{FedPseudo} performs similar to the DeepSurv model when it shows significant improvement over the CoxPH and DeepHit model on both centralized and federated settings. However, even though CoxPH performs surprisingly well on real data due to the absence of interaction effect among the covariates, absence of nonlinearity, or limited samples in a client, it fails significantly on the synthetic dataset since the datasets were generated by inducing high nonlinearity. 

\noindent \textbf{Discussion:}
The CoxPH and DeepSurv can have different baseline hazard functions for different clients and may not satisfy the proportional hazard assumption for all clients. We do not need to specify the baseline hazard function for the relative risk comparisons since the risk ratio between individuals does not depend on time. However, in this work, we compared the evaluation metrics based on the survival probabilities, which need to estimate the baseline hazard function. 
Thus, we need to make a strong assumption that a common baseline hazard function for all the clients can be computed centrally using either the entire training data or computed and shared by a client. However, making this assumption leads to the leakage of the individual's information, which may break privacy laws or data protection regulations. 
A naive approach is using the test dataset to compute baseline hazard; however, it is infeasible if we only have a few test observations. 
For the discrete model, DeepHit, we first need to specify the number of time-bins for calculating the survival probability. From the experiments, we found that the performance of the DeepHit model highly depends on the number of time-bins, which is set 365 for all datasets showing overall better performance. 
Unlike the baseline FSA approaches, our FedPseudo does not make any underlying assumption except a common maximum time horizon shared among all the clients, which can be obtained without sharing any patient information.

%% file: Conclusion.tex
Federated survival analysis (FSA) is an emerging problem in federated healthcare domain. 
In this paper, we proposed a first-of-its-kind novel pseudo value-based FSA approach, \textbf{FedPseudo}, for estimating subject-specific survival probabilities in a federated fashion in the presence of censoring.
We also proposed a novel pseudo value derivation algorithm for FL settings, which speeds up the computation of pseudo values and allow us to model and analyze the survival data from different clients without sharing data. 
We showed that our proposed approach, FedPseudo, performs similar to or better than the existing FSA approaches in both real-world data and simulated data under various censoring settings. 
For future work, we will investigate advanced Federated learning algorithms to improve robustness for non-iid health data.